\documentclass{elsart}
\usepackage[latin1]{inputenc}
\usepackage{amssymb}
\usepackage{graphicx}
\usepackage{url}
\usepackage[normalem]{ulem}
\newcommand{\R}[1]{\ensuremath{\mathbb{R}^{#1}}}

\usepackage{natbib}

\newcommand{\on}[0]{in }

\newcommand{\plain}[0]{regular } 

\newcommand{\nbfeatures}[0]{n} 

\newcommand{\methodlearning}[0]{When the optimal values for the meta parameters $\rbfnbneurons$ and $\rbfwsf$ have
been determined by the 3-fold validation procedure explained above on the learning set,
the final model is fit using all data from the learning set and the results of the model
on the test set are reported}

\newcommand{\V}[1]{{\rm Var}\left(#1\right)}					

\newcommand{\RVe}[1]{{\rm \uppercase{#1}}}				
\newcommand{\RV}[1]{{\rm \uppercase{#1}}}				

\newcommand{\sq}[1]{\left(#1\right)^2}

\newcommand{\realy}[0]{y}
\newcommand{\predy}[0]{\hat y}

\newcommand{\modelinput}[0]{x}
\newcommand{\samplesize}[0]{N}

\renewcommand{\RV}[1]{{\bf \uppercase{#1}}}             

\newcommand{\rbfnbneurons}[0]{M}
\newcommand{\rbfindexneurons}[0]{l}
\newcommand{\rbfkernel}[1]{K\!\left( #1 \right)}
\newcommand{\rbfcoeff}[0]{\lambda}
\newcommand{\rbfwidths}[0]{\sigma}
\newcommand{\rbfcenter}[0]{C}
\newcommand{\rbfwsf}[0]{W\!SF}
\newcommand{\rbfbias}[0]{b}

\renewcommand{\d}[0]{n}
\newcommand{\N}[0]{P}
\newcommand{\freev}[1]{#1} 

\newcommand{\p}[1]{\left( #1 \right)}
\newcommand{\pdf}[1]{\mu_{#1}\!}

\newcommand{\I}[1]{I\!\p{#1}}

\newcommand{\NMSE}[0]{N\!M\!SE}
\newcommand{\diffentropy}[1]{ -\int \pdf{\RVe{#1}}(\freev{#1}) \,\log\, \pdf{\RVe{#1}}(\freev{#1}) \,{\rm d}\freev{#1} }
\newcommand{\conddiffentropy}[2] {-\int \pdf{\RVe{#1},\RVe{#2}}\p{#1,#2} \,\log\, \pdf{\RV{y}}\p{\freev{#1}|\RV{#2}=\freev{#2}} d\freev{#1}d\freev{#2}}

\newcommand{\testsetsize}[0]{P_t}
\newcommand{\learningsetsize}[0]{P_l}

\begin{document}
\sloppy

\begin{frontmatter}

\title{Fast Selection of Spectral Variables with B-Spline Compression}

\author[INRIA]{Fabrice Rossi\corauthref{cor}}, %
\ead{Fabrice.Rossi@inria.fr}
\author[CESAME]{Damien Francois}, %
\ead{francois@csam.ucl.ac.be}
\author[CESAME]{Vincent Wertz}, %
\ead{wertz@inma.ucl.ac.be}
\author[BNUT]{Marc Meurens}, %
\ead{meurens@bnut.ucl.ac.be}
\and \author[DICE]{Michel Verleysen} %
\ead{verleysen@dice.ucl.ac.be}

\address[INRIA]{Projet AxIS, INRIA, Domaine de Voluceau, Rocquencourt,
  B.P. 105, 78153 Le Chesnay Cedex, France}
\address[CESAME]{Universit\'e catholique de Louvain - Machine Learning Group,
  CESAME, 4 av. G. Lema\^{\i}tre, 1348 Louvain-la-Neuve, Belgium}
\address[BNUT]{Universit\'e catholique de Louvain - Laboratory of Spectrophotometry, BNUT, Place Croix du Sud 2/8,
  1348 Louvain-la-Neuve, Belgium} 
\address[DICE]{Universit\'e catholique de Louvain - Machine Learning Group,
  DICE, 3 place du Levant, 1348 Louvain-la-Neuve, Belgium}

\corauth[cor]{Corresponding author.}

\begin{abstract}
  The large number of spectral variables in most data sets encountered in
  spectral chemometrics often renders the prediction of a dependent variable
  uneasy.  The number of variables hopefully can be reduced, by using either
  projection techniques or selection methods; the latter allow for the
  interpretation of the selected variables.  Since the optimal approach of
  testing all possible subsets of variables with the prediction model is
  intractable, an incremental selection approach using a nonparametric
  statistics is a good option, as it avoids the computationally intensive use
  of the model itself. It has two drawbacks however: the number of groups of
  variables to test is still huge, and colinearities can make the results
  unstable.  To overcome these limitations, this paper presents a method to
  select groups of spectral variables.  It consists in a forward-backward
  procedure applied to the coefficients of a B-Spline representation of the
  spectra.  The criterion used in the forward-backward procedure is the mutual
  information, allowing to find nonlinear dependencies between variables, on
  the contrary of the generally used correlation.  The spline representation
  is used to get interpretability of the results, as groups of consecutive
  spectral variables will be selected. The experiments conducted on NIR
  spectra from fescue grass and diesel fuels show that the method provides
  clearly identified groups of selected variables, making interpretation easy,
  while keeping a low computational load. The prediction performances obtained
  using the selected coefficients are higher than those obtained by the same
  method applied directly to the original variables and similar to those
  obtained using traditional models, although using significantly less
  spectral variables.
\end{abstract}


\end{frontmatter}


\section{Introduction}
Prediction problems are often encountered in analytical spectral chemometrics.
They require estimating the unknown value of a dependent variable from, for
example, a near-infrared spectrum. Such problems may be encountered in the
food \citep{Ozaki:1992}, pharmaceutical \citep{Blanco:1999} and textile \citep{Blanco:1997} industry, to cite only a few.

Viewed from a statistical or data analysis perspective, the main difficulty in
such problem is to cope with the colinearity between spectral variables: not
only consecutive variables in a spectrum are highly correlated by nature, but
in addition real applications usually concern databases with a low number of
known spectra, and a high number of spectral variables. Any method built on
the original spectral variables is thus ill-posed, making feature (spectral
variable) selection and/or projection necessary.

Selection and projection methods differ by several aspects. Projection
methods are more general by essence, as selection may be regarded as
projection with many zero weights. However, projection methods usually build
factors (latent variables) that are combinations of a large number of original
features. Even if their prediction properties are good, they usually suffer
from the fact that the latent variables are hardly interpretable in terms of
original features (wavelengths in the case of infrared spectra). On the
contrary, selection methods are based on the principle of choosing a small
number of variables among the original ones, leading to easy interpretation.
Of course, the challenge with selection methods is to obtain prediction
performances of the same level as projection ones.

In this work, we are interested in variable selection methods providing
interpretability. However, if the whole procedure consisting in selecting the
features and building a prediction model on them is kept linear, it will
certainly lead to poorer performances than the traditional and widely used
PLS (Partial Least Squares), as the latter consists in a projection and a
prediction. It is thus investigated how nonlinear models may be used, both
for selecting the features and performing the prediction.

Nonlinear models could be used in a wrapper approach \citep{Kohavi:1997}, in
which their estimated generalization performances is used as a relevance
criterion for a group of variables. This however, is very demanding in terms
of computational load because resampling techniques must be used to estimate
accurately the predicted error of the model, in addition to the fact that one
model must be learned for each considered feature set. This paper thus focuses
on the so-called filter approach: the features are selected prior the use of
any prediction model.

Among filter methods, the correlation is the standard criterion to be used for
selecting features in a linear way: features with maximal correlation with the
dependent (output) variable, and possibly with minimal information between
them to avoid redundancy, are selected. Mutual information (see e.g.
\citet{CoverThomasInformationTheory:1991}) extends the correlation to the
measure of nonlinear dependencies, while correlation is strictly limited to
linear ones. As an example, the correlation between a centered antisymmetric
variable and its second power is zero, despite the fact they obviously depend
one from another (though in a nonlinear way). The mutual information avoids
this drawback, providing a more general and less restricted way to measure
dependencies between variables.

The mutual information (MI) has already been used to select variables from
near-infrared spectra \citep{Rossi:2005}. Despite it provides a promising way
to extend state-of-the-art spectral analysis to nonlinear methodologies, the
direct selection of variables by MI suffers from some drawbacks. First, the MI
estimation becomes difficult as the number of selected variables grows.
Indeed in a forward procedure the estimation is faced to the curse of
dimensionality, making the estimation of the MI with the last selected feature
much more difficult than with the first selected one. Second, the low number
of spectra usually available for learning makes the results of the selection
highly dependent on the data set: a small change in the data can lead to
different selected variable sets, resulting in difficult interpretation.
Finally, even though the estimation of the mutual information is less
demanding in terms of computation time than the construction of a nonlinear
model, the large number of initial variables results in high computation times
for the selection.

In this paper, we propose to first reduce the number of variables through a
projection of the spectral features before the selection by mutual
information. To maintain the interpretability despite the use of a projection,
the latter is achieved by ensuring that each coordinate in the projection
corresponds to a restricted set of initial features with consecutive
wavelengths. The general methodology proposed in
\citep{Alsberg1993Representation} is followed: spectra are projected on a
functional basis. More precisely, as in e.g.
\citep{AlsbergKvalheim1993Bsplines}, a projection on a basis of B-splines is
chosen, rather than wavelets for example; indeed B-splines have the advantage
that they span a restricted interval of wavelengths, and that the intervals
are roughly of the same length over the whole range. As a consequence, each
coefficient depends on the value of the corresponding spectrum on a limited
wavelength interval. The complete procedure then consists in replacing the
spectra by their B-spline coefficients, in selecting relevant coefficients by
measuring their mutual information with the output variable, and by predicting
the latter using Radial-Basis Function networks (any other nonlinear model
could be used). All three steps are nonlinear, giving to the procedure the
necessary flexibility to reach high performances both in prediction and in
interpretation. Design parameters that are unavoidable in a nonlinear context,
such as the number of B-splines to be used in the projection, are set
automatically (without the necessity of a user's choice) using a
cross-validation method. 

This paper shows that the prediction results obtained by this procedure are
comparable than those obtained through conventional linear
techniques such as PLS. In addition interpretability is added, as the number
of wavelengths selected by the procedure remains low, making it possible to
identify which wavelengths are responsible for the phenomenon to
predict. Moreover, B-spline compression allows us both to reduce the feature
selection running time and to increase the quality of the prediction results
compared to the same nonlinear procedure applied directly to the original
spectral variables. 

Section \ref{section:Bsplines} of this paper reminds how spectra can be
projected on a basis of B-splines, details how the number of B-splines can be
set automatically and analyzes the computational complexity of the procedure.
Section \ref{section:Selection} presents the mutual information criterion and
its use in a forward-backward procedure. It also investigates the computational
complexity of the forward-backward method.  Section \ref{section:Results}
shows examples of the application of the proposed method on two data sets. The
first one consists of NIR spectra obtained from fescue grass; the aim is to
predict the nitrogen content of the plant. The second one is a database of
spectra from fuel samples for which the goal is to predict the Cetane Number of
the fuel. 

\section{B-splines}\label{section:Bsplines}
\subsection{Functional representation of spectra}
As pointed out in the introduction, the performances of variable selection
procedures decrease with the number of initial variables, while their running
time increases. Our goal is therefore to reduce the set of initial variables
in a simple way that preserves information and interpretability. This can be
done by leveraging the functional nature of spectra, following the general
approach initiated in \citep{Alsberg1993Representation} and the principals of
Functional Data Analysis \citep{Ramsay:1997}. 

A spectrum can be viewed as a smooth function $s$ that maps a wavelength
interval $[w_{\min},w_{\max}]$ to the measured response, for instance the
transmittance of the studied sample. A spectral variable $X_w$ corresponds to
the value taken by the function at a specific wavelength, i.e., to
$X_w(s)=s(w)$ for a given wavelength $w\in[w_{\min},w_{\max}]$. We denote
$w_1,\ldots,w_N$ the wavelengths used by the spectrometer (numbered in
increasing order) and $X_1,\ldots,X_N$ the corresponding original spectral
variables. 

A simple way to reduce the number of variables is to replace each spectrum by
its best approximation by a linear combination of $n$ basis functions, with
$n$ smaller than $N$. Let us consider $n$ functions $(\phi_i)_{1\leq i\leq n}$
from $[w_{\min},w_{\max}]$ to \R{} (the set of real numbers). The best
approximation of a spectrum $s$, in the sense of the squared reconstruction
error, is obtained by minimizing the following error with respect to
$(a_{s,i})_{1\leq i\leq n}$:
\begin{equation}
  \label{eq:SquareReconstructionError}
\sum_{j=1}^N\left(s(w_j)-\sum_{i=1}a_{s,i}\phi_i(w_j)\right)^2.
\end{equation}
This type of quadratic optimization problem is easy to solve; it is well
known, see e.g. \citep{Ramsay:1997}, that the $(a_{s,i})_{1\leq i\leq n}$ are
obtained from the $(s(w_j))_{1\leq j\leq N}$ via a linear transformation that
depends only on the $(\phi_i)_{1\leq i\leq n}$ and on the $(w_j)_{1\leq j\leq
  N}$. In other words, there is a $n\times N$ matrix $R$ such that for all $s$
and $i$:
\begin{equation}
  \label{eq:NewVariablesFromOldVariables}
a_{s,i}=\sum_{j=1}^NR_{i,j}s(w_j). 
\end{equation}
This allows us to define $n$ new variables $A_1,\ldots, A_n$ from the $N$
original ones by:
\begin{equation}
  \label{eq:NewVariablesFromOldVariablesTwo}
A_i=\sum_{j=1}^NR_{i,j}X_j.
\end{equation}
The main difficulty of this approach lies in the choice of the set of
functions $(\phi_i)_{1\leq i\leq n}$. They must provide good
approximations of the original spectra for a small value of $n$ (compared to
$N$) while preserving interpretation abilities: in practice, we need each
$A_i$ to depend only on a small localized subset of the original variables. 

\subsection{Spline approximation}
A simple solution is obtained by using a spline approximation, i.e. a
piecewise polynomial representation of the spectra. This is done by splitting
the original wavelength range into $p$ sub-intervals, defined by the $p+1$
values, $t_0,\ldots,t_p$, called \emph{knots}, such that $t_{i}<t_{i+1}$,
$t_0=w_{\min}$ and $t_{p}=w_{max}$. A spline of order $d$
\citep{DeBoorBookSpline} is a function $f$ from $[w_{\min},w_{\max}]$ to \R{}
such that:
\begin{itemize}
\item $f$ is a polynomial of degree $d-1$ on each interval $[t_{k},t_{k+1}[$;
\item $f$ is $C^{d-2}$ on $[w_{\min},w_{\max}]$ (i.e., $f$
is continuous and has continuous derivatives up to order $d-2$).
\end{itemize}
The regularity constraints imposed on splines are adapted to spectrometric
applications: spectra are generally very smooth and are therefore very
accurately represented by splines of small order (e.g. $4$ or $5$). 

The vector space of splines of order $d$ based on the knots $(t_k)_{0\leq
  k\leq p}$ has a basis made of $p-1+d$ \emph{B-splines} (see e.g.
\citep{DeBoorBookSpline} for details), $B^d_1,\ldots,B^d_{p-1+d}$. Each
B-spline is a spline with a localized support: it is positive on only at most
$d$ consecutive intervals. This basis can be used to define $n=p-1+d$ new
variables, as proposed in the previous section. It should be noted that
choosing $p$ and $d$ is not enough to define an unique B-spline basis: the
positions of the knots have to be specified. In this paper, we split
$[w_{\min},w_{\max}]$ into $p$ sub-intervals of equal length, but adaptive
schemes could be used as long as the knots are identical for all spectra. 

B-splines are very computational efficient. In the case of arbitrary functions
$(\phi_i)_{1\leq i\leq n}$, calculating $(a_{s,i})_{1\leq i\leq n}$ for one
spectrum costs $O(n^2N)$ operations (when $n\leq N$), whereas its is only
$O(N)$ for $n$ B-splines, because of the localized supports (see e.g.
\citet{Ramsay:1997}). Computational details on B-splines can be found in
\citet{DeBoorBookSpline} or in e.g. 
\citet{AlsbergKvalheim1993Bsplines,OlssonEtAl1996Bsplines}.

\subsection{B-spline coordinates}\label{section:Bsplines:coordinates}
Spectra representation by B-splines of degree 0 (i.e. of order 1), which
corresponds to piecewise constant approximation, has been used for compression
purpose, see, e.g.,
\citet{AlsbergKvalheim1993Bsplines,AlsbergEtAl1994Bsplines,AlsbergKvalheim1994ThreeMode,OlssonEtAl1996Bsplines}.
Those papers take advantage of the linear relationship between a spectrum and
its coordinates on the B-spline basis. When the coordinates are used as inputs
to a linear method (such as Principal Component Analysis, as in
\citet{AlsbergKvalheim1994ThreeMode}), the results of the method can be
applied directly to the original spectra by combining both linear
transformations. In our application, we use the compression property already
explored in earlier work, but we also take advantage of the localization
properties of the B-splines to preserve interpretability of the variables.

\begin{figure}[htbp]
  \centering
  \includegraphics[angle=270,width=\textwidth]{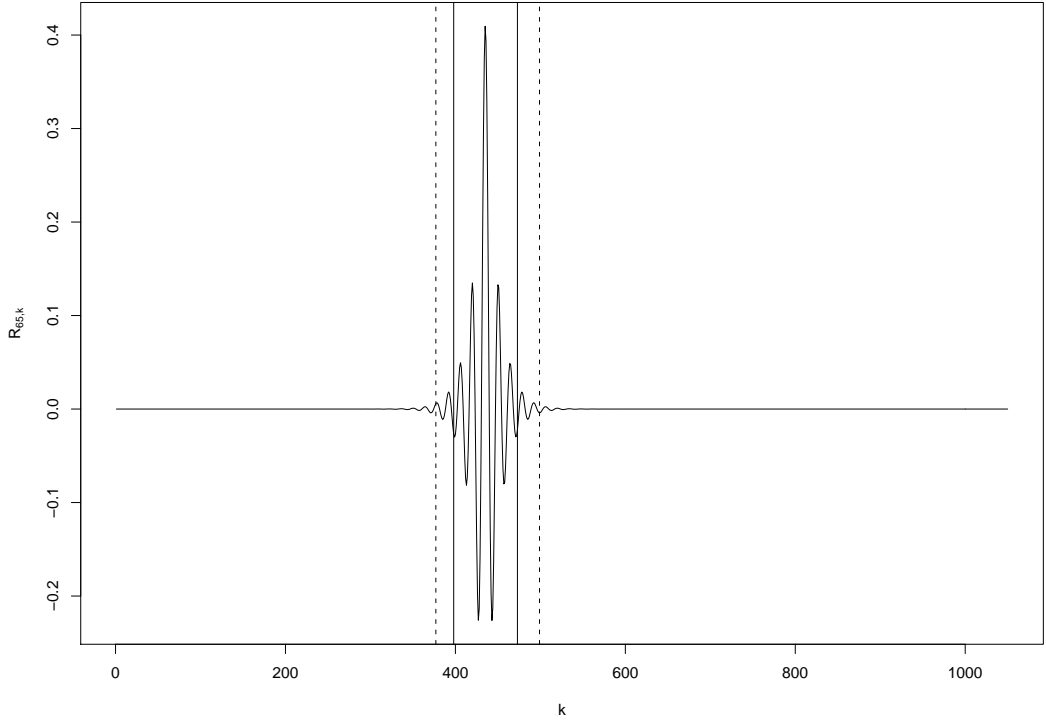}
  \caption{Graphical representation of $R_{65,k}$ for a B-spline of order 5
    basis with 155 B-splines calculated for 1050 original spectral variables}
  \label{fig:BsplineCoefficient}
\end{figure}

Because of the localized supports of the B-splines, the linear transformation
expressed by the matrix $R$ used in Equations
\ref{eq:NewVariablesFromOldVariables} and
\ref{eq:NewVariablesFromOldVariablesTwo} has an interesting property: most of
the values in $R$ are very small and significant values are localized, as
illustrated \on Figure \ref{fig:BsplineCoefficient}. This property preserves
interpretability as it allows to determine which wavelength range of the
original spectra contributes significantly to the value of a chosen new
variable.

Based on this idea, a wavelength range $[w_{l_i},w_{u_i}]$ for $A_i$ can be
easily estimated. Given a precision ratio $\epsilon>0$, the
indexes of the bounds of the interval are
\begin{eqnarray}
  \label{eq:BsplineRange}
l_i&=&\max\left\{1\leq j\leq N\,\Bigg|
  \max_{1\leq k<j}|R_{ik}|<\epsilon\max_{1\leq k\leq N}|R_{ik}|  \right\},\\
u_i&=&\min\left\{1\leq j\leq N\,\Bigg|
  \max_{j< k\leq N}|R_{ik}|<\epsilon\max_{1\leq k\leq N}|R_{ik}|  \right\},
\end{eqnarray}
with the convention that $\max_{1\leq k<1}|R_{ik}|=\max_{N< k\leq
  N}|R_{ik}|=0$. The lower bound $w_{l_i}$ corresponds to the largest index
$j$ such that all coefficients $R_{ik}$ for $k<j$ are smaller than $\epsilon$
times the maximal coefficient. The upper bound $w_{u_i}$ is defined in a
symmetric way. Figure \ref{fig:BsplineCoefficient} displays two wavelength
intervals: the vertical solid lines give the bounds of the interval calculated
for $\epsilon=0.05$ and the dashed lines correspond to $\epsilon=0.01$.

\subsection{Choosing the B-spline basis}\label{sec:bsplineloo}
A critical point of this approach is to determine a correct value for $n$, the
number of new variables. A small value of $n$ corresponds to an efficient
variable selection but also to a poor approximation of the spectra. On
the contrary, a high value for $n$ ensures almost perfect approximation of the
spectra, but does not bring any improvement in terms of variable selection.

For efficiency reasons, a wrapper approach, in which the optimal value of $n$
would be chosen according to the quality of prediction model built with the
selected variables, is not possible; a filter approach is preferred. However,
the \plain approximation error of the spline representation is not a reliable
criterion: it will tend to favors the highest possible value of $n$, i.e.
almost $N$.

Our solution is based on a leave-one-out (loo) criterion (as in
\citet{RossiEtAl05Neurocomputing}). Let us first recall the definition of
the leave-one-out error for a single spectrum $s$ and a specific B-spline
basis $B^d_1,\ldots,B^d_{n}$. The loo error is based on estimating the effects
of removing one evaluation point of $s$ (a wavelength) on the quality of the
approximation of $s$ at this point.  To do this, we define for all $1\leq
k\leq N$ the coordinates $a_{s,i}^{(-k)}$ as the optimal coefficients for the
B-spline representation of
$(s(w_1),\ldots,s(w_{k-1}),s(w_{k+1}),\ldots,s(w_{N}))$. Those coefficients
minimize
\begin{equation}
  \label{eq:CoordinatesLoo}
\sum_{j=1,j\neq k}^N\left(s(w_j)-\sum_{i=1}^{n}a^{(-k)}_{s,i}B^d_i(w_j)\right)^2.
\end{equation}
The leave-one-out error for the spectrum $s$ is then
\begin{equation}
  \label{eq:LooOneSpectrum}
LOO(s,n)=\frac{1}{N}\sum_{k=1}^N\left(s(w_k)-\sum_{i=1}^{n}a^{(-k)}_{s,i}B^d_i(w_k)\right)^2.
\end{equation}
The advantage of the loo error over the \plain approximation error is that it
favors stable solutions for which the removal of one observation does not
modify significantly the spline approximation.

The calculation of the loo error might appear 
computationally-intensive, but efficient algorithms exist (see e.g.
\cite{Ramsay:1997}): they scale in $O(nN)$ for one spectrum.

The total leave-one-out error for a set of $P$ spectra is obtained by simply
summing the individual loo errors, i.e. $LOO(n)=\sum_{l=1}^PLOO(s_l,n)$ (the
computation cost is $O(nNP)$). This value is used to select the optimal $n$,
via a simple brute force minimization. The simplest solution consists in
calculating $LOO(n)$ for all possible values of $n$ in a reasonable range, for
instance $[N/20,N/2]$. The worst case total complexity of this approach is
dominated by $O(N^3P)$. Heuristics can be used in practice to reduce this
cost, for instance by testing a few values in $[N/20,N/2]$ and then by testing
all the possible solutions in a small sub-interval of $[N/20,N/2]$.

\section{Variable selection}\label{section:Selection}

The projection of each spectrum as described in the previous section results in $n$ B-spline coefficients.  The next objective is to select which of these coefficients are important for the prediction of the output (response) variable.  This is achieved through variable selection.

The benefits of variable selection, or feature selection are twofold. First, it allows building an efficient prediction model of the response variable, which we call $\RV{y}$,  by reducing the data space dimension.  As the complexity of most model structure increases at best linearly, and at worst exponentially with the dimension of the data, reducing the dimension can help avoiding overfitting and reducing computation times. Second, feature selection methods identify which features are relevant for the problem at hand. Although they do not help discovering the mechanisms by which the inputs interact together, they can identify the elements that have an influence on the problem considered. This provides the interpretability that is needed in most real-world applications.  In this section, a set of features will be denoted $\RV{x}$ and contains either spectral variables $X_w$ or spline coordinates $A_i$. It is viewed here as random vector whose dimension $\d$ is the number of spectral or spline variables it contains.

A feature selection method needs to combine two elements. The first one is a
measure to score a feature subset to evaluate its potential for
prediction. The mutual information criterion is a good choice as it is
nonparametric (it does not assume any distribution of the features) and
model-independent (it is generic and does not make use of a specific
prediction model).  The second element is a procedure that explores the
candidate feature subset space in order to find the optimal one. This can be
done with incremental algorithms such as the forward-backward search
procedure. Although this class of algorithms is sub-optimal in the sense that
it does not ensure finding the optimal feature subset among all possible ones,
it often presents a good trade-off between accuracy and computation time.

The remaining of this section will introduce the mutual information criterion and the forward-backward incremental search.

\subsection{Mutual information}\label{sec:mutualinformation}

To evaluate the relevance of a group of features in terms of prediction potential, we will estimate the mutual information between that group of features and the variable to predict.  The mutual information of two variables is the amount of uncertainty that is lost on one variable when the other is known, and \emph{vice versa}.

The uncertainty of a variable can be estimated through its entropy. The entropy of a real-valued random vector $\RV{y}$ is a non-negative value given by \citep{Shannon:1948}
\begin{equation}
H(\RV{y}) = \diffentropy{y},
\end{equation}
where $\pdf{\RV{y}}$ is the probability density function of $\RV{y}$.

The entropy of $\RV{y}$ when the value of some other random vector $\RV{x}$ is known is the \emph{conditional entropy}:
\begin{equation}
H(\RV{y}|\RV{x}) = \conddiffentropy{y}{x}.
\end{equation}

The difference between those two values, i.e. the difference between the entropy of $\RV{y}$ and the entropy of $\RV{y}$ conditioned on $\RV{x}$ is called the mutual information between $\RV{y}$ and $\RV{x}$:
\begin{equation}
\I{\RV{y},\RV{x}} = H(\RV{y}) - H(\RV{y}|\RV{x}).
\end{equation}
It is symmetric and measures the amount of information a variable can bring about the other.
The mutual information is zero if and only if the variables are independent; it is thus well suited to measure the relevance of $\RV{x}$ to predict the values of $\RV{y}$ \citep{Battiti:1994}.

In practice, the mutual information has to be estimated from the data set, as the exact probability density functions in the above equations are not known. The most sensitive part of the estimation of the mutual information is the estimation of the joint probability density function $\pdf{\RV{y},\RV{x}}$. Several methods have been developed in the literature to estimate such joint densities \citep{Scott:1992}. Unfortunately, most of them require a sample whose size grows exponentially with both the dimensions of $\RV{x}$ and $\RV{y}$ to provide an accurate estimation. Since most applications consider one output at a time, the dimension of $\RV{y}$ is one. However, in the next section, we will see that the mutual information has to be estimated between a set of features and the variable to predict; therefore, the dimension of $\RV{x}$ grows and can potentially be as large as the total number of features.

A method that does not so dramatically depend on the sample size is the method developed by \citep{Kraskov:2004}. It is based on a nearest neighbors statistic. The core of the algorithm is the assumption that data element that are close in the space will correspond to similar values of the variable to predict.

The algorithm described in \citep{Kraskov:2004,Rossi:2005} needs $O(\d \N^2)$
operations, where $\d$ is the dimension of $\RV{x}$ and $\N$ is the sample
size (here the number of spectra). However, using heuristics, the algorithm
has been implemented in such a way to have an average complexity that is
linear in both the dimension and the sample size\footnote{The MILCA toolbox:
  \url{http://www.fz-juelich.de/nic/cs/software/}}.

\subsection{Forward-backward procedure}\label{eq:fb}

Searching for the optimal (according to the mutual information criterion for example) feature subset actually requires to evaluate the mutual information between all $2^\d-1$ possible subsets and the variable to predict.  This, especially for spectrophotometric data, is often intractable. Combinatorial optimization algorithms, such as a genetic or simulated annealing ones, are rather efficient and could be used (see Kohavi and John 1997); they however demand a lot of computations to converge.  Incremental (greedy) algorithms are cheaper and usually perform efficiently too \citep{Aha:1996}.  They are suboptimal in the sense that there is no guarantee that they will find the optimal subset, because they choose one feature at a time and never question that choice afterward. However, they only need $O(\d^2)$ evaluations of the mutual information to find the (sub-optimal) solution.  They nevertheless will often find a good subset. The procedure actually is optimal when the features are independent.

The forward-backward procedure acts in two stages.  The first stage, the forward phase, consists in adding features one  by one. At each iteration, the feature chosen to incorporate the current subset is the one that most increases the mutual information with the variable to predict.  The process is stopped when adding any new feature actually decreases the mutual information.  The second stage is the backward phase.  During this stage, features are eliminated one at a time.  The feature that is excluded from the current feature subset is the feature that most increases the mutual information when it is discarded. As in the first stage, the backward phase stops when discarding any other feature decreases the mutual information of the subset with the variable to predict.

\subsection{Computational cost}
Let us consider $\nbfeatures$ initial variables. At each iteration of the forward stage, all features that are not already selected in the current set have to be tested.  So at iteration $i$ there are $\d-i$ subset evaluations to perform. Since there are $\d$ features at the beginning, the maximum number of evaluations is:
\begin{equation}
\sum_{i=0}^{\d-1} \d-i = \frac{3\d}{2} (\d+1) .
\end{equation}

Since the nearest-neighbor-based estimation of the mutual information is linear in the dimension $\nbfeatures$ and quadratic in the number of data points $P$, the total complexity of the algorithm is O($\nbfeatures^3P^2$). Similar calculations lead to the same result for the backward procedure.

Consequently, reducing in advance the initial number of features can dramatically decrease the computation time. Of course, the method used to reduce the initial number of features must not be more complex than the forward-backward procedure.

Given the complexity of the loo estimation for the splines detailed in Section
2.3, and the fact that at most O($N$) B-splines can be used, the total worst
case complexity of the spline construction procedure is O($N^3P$).

If the forward-backward method is applied to the $N$ spectra variables, it will of course require O($N^3P^2$) operations.

Therefore, since the value of $n$ determined by leave-one-out is often much smaller than $N$, the overall cost of the procedure, that is O($N^3P$) + O($n^3P^2$) is smaller than the cost of the forward-backward procedure conducted on all the spectral variables, which is O($N^3P^2$). Actually, working with the spline coordinates instead of the spectral variables is much cheaper when
\begin{equation}
\frac{1}{P} + \left( \frac{n}{N} \right)^3 < 1,
\end{equation}
which is satisfied most of the time in practice.

\section{Experimental results}\label{section:Results}
The selection of variables with the mutual information criterion, after projection of the spectra on a basis of B-splines, leads to a prediction model that combines advantages in terms of performances and interpretability.  In this section, the prediction methodology is first summarized, before describing the data sets on which experimental comparisons are performed.

\subsection{Methodology}
In order to asses the propositions contained in this paper, the proposed model
is compared to reference ones; the models and comparison criteria are detailed
in the following sections.

\subsubsection{Proposed method}
To achieve a good prediction of the output in a reasonable time as well as an
interpretable determination of the wavelengths involved in the process, our
method consists in the following steps:
\begin{enumerate}
\item Extraction of the B-spline coefficients for each spectrum. The number of
  B-splines is chosen by the leave-one-out procedure described in Section
  \ref{sec:bsplineloo}.
\item Selection of the B-spline coefficients through mutual information
  maximization (Section \ref{sec:mutualinformation}) and forward-backward
  search (Section \ref{eq:fb}).
\item Calculation of the wavelength ranges associated to the selected
  variables, as explained in Section \ref{section:Bsplines:coordinates}, with
$\epsilon=0.01$.
\item Construction of a nonlinear model (Radial Basis Function Network) on the
  coefficients selected by the previous step.
\end{enumerate}
A Radial Basis Function Network (RBFN) model is a weighted sum of Gaussian
kernels \citep{Powell:1987}. The prediction $\predy$ of $\realy$ given
$x$ is computed as

\begin{equation}
\predy = \sum_{\rbfindexneurons = 1}^\rbfnbneurons \rbfcoeff_\rbfindexneurons \cdot \rbfkernel{\modelinput, \rbfcenter_\rbfindexneurons, \rbfwidths_\rbfindexneurons} + \rbfbias,
\end{equation}
where
\begin{equation}
\rbfkernel{\modelinput, \rbfcenter_\rbfindexneurons, \rbfwidths_\rbfindexneurons} = \exp\left({-\sq{\frac{\|\modelinput - \rbfcenter_\rbfindexneurons\|}{\rbfwsf\cdot\rbfwidths_\rbfindexneurons}}}\right).
\end{equation}
The position of the centers $\rbfcenter_\rbfindexneurons$ is determined by
vector quantization \citep{Powell:1987}, and the values of
$\rbfwidths_\rbfindexneurons^2$ are set to the variance of the clusters
identified by the vector quantization stage.  The $\rbfcoeff_\rbfindexneurons$
and $\rbfbias$ are fixed by linear regression, as described in
\citep{Benoudjit:2004}.

Both $\rbfnbneurons$ and $\rbfwsf$ are called meta parameters of the model
and determine its complexity,  hence its generalization capabilities. They
are chosen so as to minimize the generalization error of the model.

The generalization error is estimated using a 3-fold cross validation
technique.  The learning set is split into three different sets of equal size.
Each subset serves as a validation set, one at a time, while the other two sets are
used to build the model.  Although the population of each set is randomly
chosen, the split is done so as to ensure that each set is representative of
the distribution of  the variable $\RV{y}$ to predict.

\subsubsection{Reference methods}

In order to assess the performances of the proposed method, its results are
compared to the ones obtained by four different reference methods:
\begin{itemize}
\item to show the interest of the B-spline compression, we apply the variable
  selection method described in Section \ref{section:Selection} directly to
  the original spectral variables (this is a simplified version of the
  variable selection method proposed in \citet{Rossi:2005}). We build a RBFN
  on the selected variables, using the same procedure as the one used for the
  proposed method;
\item to motivate the use of a nonlinear model, we also include the results
of a standard linear regression (LR) built on the variables selected by the
proposed method;
\item we use linear reference models, namely a principal component
regression (PCR) and a partial least squares regression (PLSR). The numbers of
components in the PLS and in the PCR model are chosen with the same 3-fold
cross-validation method used to choose the meta parameters of the nonlinear
model.
\end{itemize}
The comparison of the models is done according to the Normalized Mean Squared
Error (NMSE) they reach on an independent test set. The test set contains
$\testsetsize$ spectra that were not used to build the models; the remaining
$\learningsetsize$ spectra are used to design the model; they form the learning
set. \methodlearning.

The NMSE normalizes the mean squared error by the variance of the output:
\begin{equation}
\NMSE = \frac{1}{{\V{\RV{y}}}}\,\sum_{y\in \textrm{test set}}^{\samplesize} \left(\realy-\predy\right)^2 .
\end{equation}
where $\predy$ is the approximation of $\realy$ by the model.  The variance is
estimated over the union of the learning and the test set.

Finally, we use a simple method to extract the wavelengths that play a
significant role in the prediction of the target variable by the best linear
model obtained with PCR or PLSR. The output of such a model can be written
\begin{equation}
  \label{eq:LinearModel}
  \predy=\alpha_0+\sum_{i=1}^{N}\alpha_iX^s_{w_i},
\end{equation}
where $X^s_{w_i}$ is a scaled version of the original input variable
$X_{w_i}$ (i.e., $X^s_{w_i}$ has zero mean and unit variance). As in section
\ref{section:Bsplines:coordinates}, we consider that wavelength $w_i$ is
important if $|\alpha_i|>\epsilon\max_{1\leq k\leq N}|\alpha_k|$.

\subsection{Data sets}\label{subsection:Datasets}
The experiments are conducted on two different databases. The first one
consists in spectra of fescue grass (shootout database) and the second one is a
data set of NIR spectra of diesel fuels (diesel database).

\begin{figure}[htbp]
  \centering
  \includegraphics[angle=270,width=\textwidth]{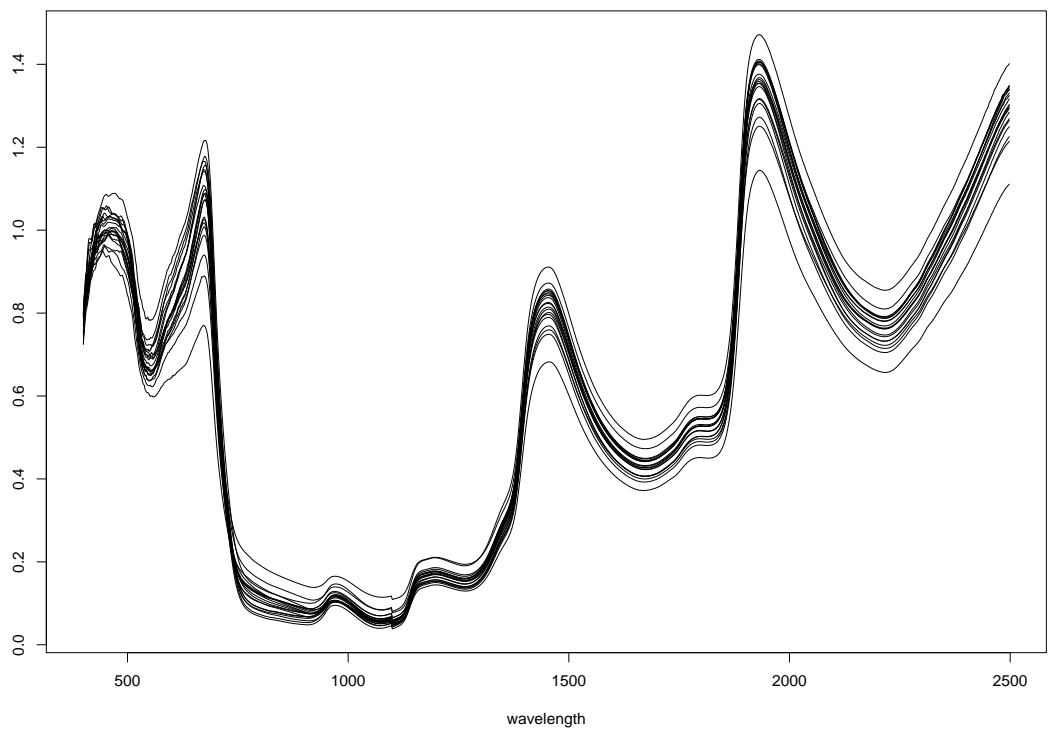}
  \caption{Some spectra from the shootout database}
  \label{fig:ShootoutSpectra}
\end{figure}

The shootout database (see Figure \ref{fig:ShootoutSpectra}) originates from a
software contest organized at the International Diffuse Reflectance
Conference\footnote{\url{http://www.idrc-chambersburg.org/index.htm}} held in
1998 in Chambersburg, Pennsylvania, USA. It consists of scans and chemistry
gathered from fescue grass (\emph{Festuca elatior}).  The grass was bred on
soil medium with several nitrogen fertilization levels. The aim of the
experiments was to try to find the optimum fertilization level to maximize
production and minimize the consequences on the environment. In this context,
the problem to address is the following: can NIR spectrometry measure the
nitrogen content of the plants?

Although the scans were performed on both wet and dry grass samples, we only
consider wet samples here (i.e. the scans were performed directly after
harvesting). The data set contains 141 spectra (see Figure
\ref{fig:ShootoutSpectra} for 20 of them) discretized to 1050 different
wavelengths, from 400 to 2 498 nm. The nitrogen level goes from 0.8 to 1.7
approximately. The data can be obtained from the Analytical Spectroscopy
Research Group of the University of
Kentucky\footnote{\url{http://kerouac.pharm.uky.edu/asrg/cnirs/shoot_out_1998/}}.

We have split randomly the data set into a test set containing 36 spectra and a
training set with the remaining 105 spectra. The random split has been done in
a way that roughly preserves the distribution of the target variable (the
nitrogen level).

\begin{figure}[htbp]
  \centering
  \includegraphics[angle=270,width=\textwidth]{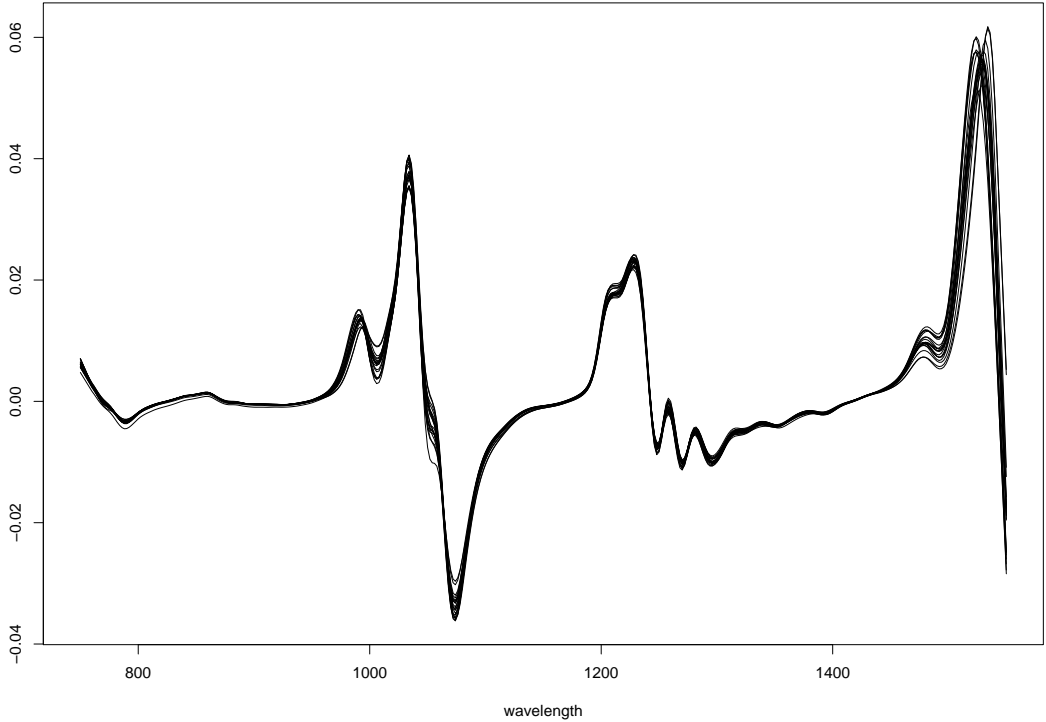}
  \caption{High leverage spectra (after centering and reduction) from the
    Diesel database}
  \label{fig:DieselSpectra}
\end{figure}

The diesel database (see Figure \ref{fig:DieselSpectra}) was built by the
Southwest Research Institute under a U.S.  Army
contract\footnote{\url{http://www.swri.org/}} (it can be obtained from
Eigenvector Research
Incorporated\footnote{\url{http://software.eigenvector.com/Data/SWRI/}}).  It
consists of scans of approximately 250 diesel fuel samples. The research was
conducted to develop instrumentation for fuel quality assessment on battle
fields. The aim was to predict several quantities from the NIR analysis:
density, total aromatics, kinematic viscosity, net heat of combustion,
freezing temperature, cetane number, etc.

The database contains only summer fuels, and outliers were removed. We
consider one of the most difficult prediction tasks of the set: to predict the
cetane number (CN) of the fuel (ranging from 40 to 60). The corresponding data
set contains 20 high leverage spectra (see Figure \ref{fig:DieselSpectra}) and
225 low leverage spectra, the latter being separated into two subsets labeled
\texttt{a} and \texttt{b}. As suggested by the providers of the data, we have
built a training set with the high leverage spectra and subset \texttt{a} of
the low leverage spectra (this corresponds to 133 spectra). The test set is
made of the low leverage spectra of subset \texttt{b} (it contains 112
spectra). All spectra range from 750 to 1550 nm, discretized into 401
wavelength values.

\subsection{Results}
The methodology proposed in this paper and the reference methods are applied
on the two databases described in the previous section.  The results are
expressed in terms of Normalized Mean Squared Error (NMSE) with normalization
variance calculated on the whole data set (learning and test).

The experiments have been conducted with Matlab. Mutual information
calculation is performed with MILCA\footnote{The MILCA toolbox:
  \url{http://www.fz-juelich.de/nic/cs/software/}} (written in C). On the
computation time point of view, it has been experimentally verified that
running the variable selection method on the original spectral variables is
one order of magnitude above running the same procedure on the B-spline
coefficients. The selection of the optimal number of B-splines is of the same
order of magnitude as the latter: both are measured in minutes on a standard
personal computer, whereas running the forward backward procedure on the
original variables takes several hours. Fitting the nonlinear model on the
selected variables takes a negligible time. 

\subsubsection{Shootout database}
The shootout database is quite challenging in terms of compression as spectra
are described by 1050 variables. The leave-one-out error calculation leads to
the selection of an optimal basis of 149 B-splines of order 5 (the optimal
number of B-splines is chosen in $[50,500]$). 

The results on the test set (NMSE) for the studied methods are given in Table
\ref{table:nitrogen}. The mutual information based selection method on the
original spectra variables keeps only three of them: wavelengths 410, 414 and
720 nm (in the visible band).  However, the performances of the nonlinear
model constructed on those variables are quite low, especially compared to the
results obtained by the proposed method. Therefore, the interpretation ability
is less interesting than with the proposed method.

The 10 B-spline variables selected by maximization of the mutual information
cannot be used to construct a linear model with performances comparable to the
ones of the optimal linear models. On this problem, the nonlinear model
constructed on those variables shows clearly the best performances (reflection
spectroscopy, used in the shootout database, has frequently some nonlinear
aspects).  

\begin{table}[htbp]
  \centering
  \begin{tabular}{|l|c|c|}\hline
Method    & Variables & NMSE (test) \\\hline
PCR       & 10 & $1.57\, 10^{-1}$\\
PLSR      & 9  & $1.51\, 10^{-1}$\\
MI + RBFN & 3 & $3.91\, 10^{-1}$\\
B-Splines + MI + RBFN & 10 & $1.21\, 10^{-1}$\\
B-Splines + MI + LR   & 10 & $2.59\, 10^{-1}$\\\hline
  \end{tabular}
  \caption{Normalized mean squared error on the test set for the nitrogen
    content prediction problem (shootout database)}
  \label{table:nitrogen}
\end{table}

\begin{figure}[htbp]
  \centering
  \includegraphics[angle=270,width=\textwidth]{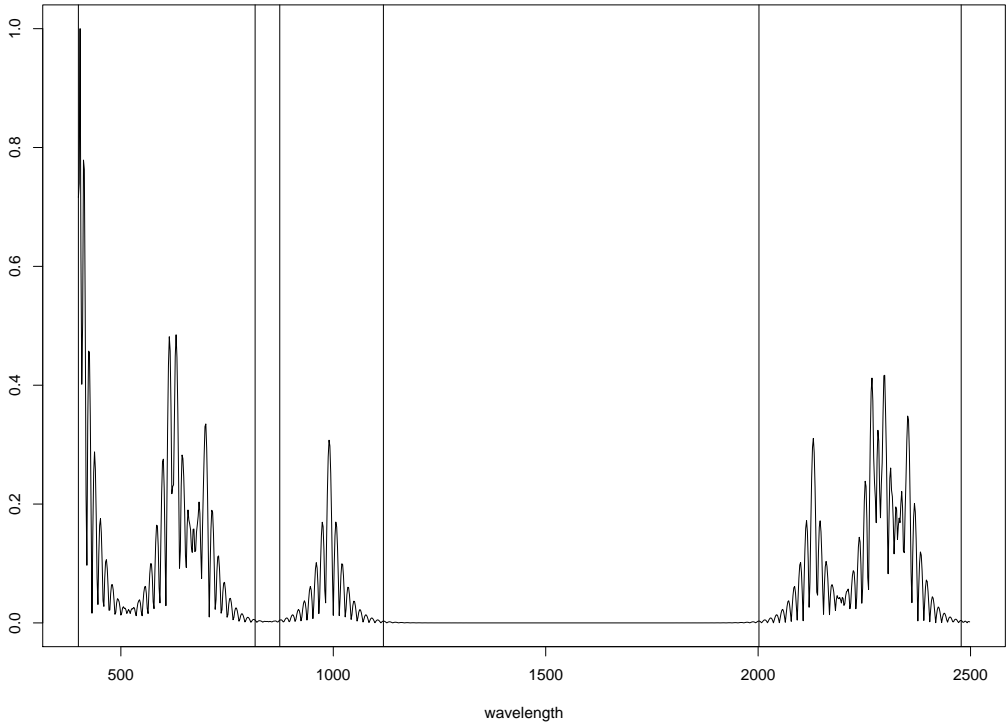}
  \caption{Normalized absolute value of the coefficients used to compute the selected variables from the original spectral variables}
  \label{fig:Nitrogen:Bsplines}
\end{figure}

While the mutual information maximization on the B-spline coefficients leads
to the selection of 10 variables, the latter correspond to only three
intervals of the original wavelength range: $[400,816]$, $[874,1118]$ and
$[2002,2478]$. Figure \ref{fig:Nitrogen:Bsplines} represents the normalized
coefficients used to compute the new variables. It appears clearly that only
some of the original wavelengths are used.

The first wavelength range corresponds to the visible band (400 to 700 nm):
this is natural as the color hue of the grass samples should be related to
their nitrogen content. The last wavelength range is also related to
absorption bands of nitrogen-hydrogen bonds. This wavelength range is not
selected when we use the feature selection algorithm on the original variable
(which retains only wavelengths in the visible range). The low performances of
this alternate solution shows, as expected, that the visible spectrum is not
sufficient to predict the nitrogen content of the sample. 

\begin{figure}[htbp]
  \centering
  \includegraphics[angle=270,width=\textwidth]{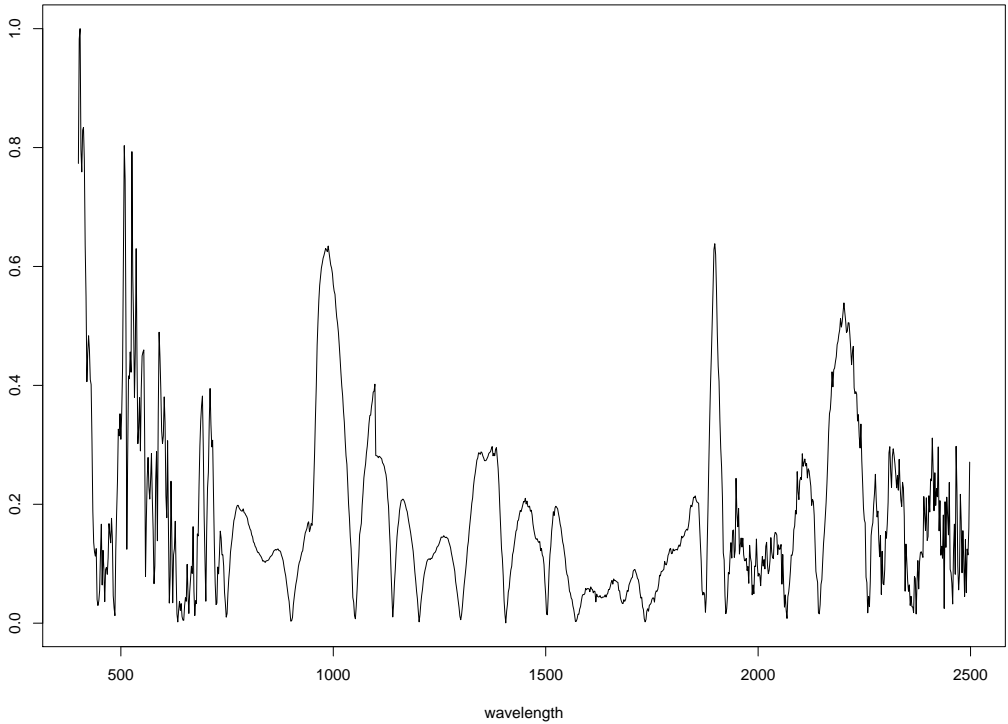}
  \caption{Normalized absolute values of the coefficients of the linear model
    induced by PLSR}
  \label{fig:Nitrogen:Linear}
\end{figure}

It is not possible to select a few wavelength ranges from the linear model
induced by the PLSR: only 17 weights out of 1050 are smaller than
$\epsilon=0.01$ times the higher one in this linear model. As illustrated in
Figure \ref{fig:Nitrogen:Linear}, the PLSR uses almost the full wavelength
range. While the PLSR model (as well as the PCR one) gives acceptable
predictions, no interpretation can be done: it seems that the model needs the
full wavelength range to provide a value for the nitrogen content. 

On this database, the proposed method allows us to obtain the best
performances, to provide interpretability, and to reduce significantly the
running time of the algorithm compared to a method where the projection on a
basis of B-splines is not used.

\subsubsection{Diesel database}
The second database corresponds to a less favorable setting for the proposed
method: spectra are obtained by transmission spectroscopy that generally leads
to linear relationship between the spectra and the target variable. Moreover,
the use of less spectral variables (401) lowers the \emph{a priori}
compression possibilities. In fact, the leave-one-out procedure selects an
optimal basis of 135 B-splines of order four (the number of B-splines is
chosen in the interval $[20,200]$). This corresponds to a reduction of
approximately one third of the number of variables: the compression ratio is
twice less important than in the case of the shootout database. 

Results on the test set (NMSE) are given in Table \ref{table:diesel:cn}. As
expected, the results of the different models are quite similar, leading to
the conclusion that the link between the cetane number and the spectrum is
linear. Moreover, all models use a rather low number of features. The
B-splines variable selected by mutual information corresponds to only three
wavelength intervals: $[816,902]$, $[954,1102]$ and $[1288,1370]$. The twelve
original variables selected by the direct procedure correspond to the
following wavelengths: 792, 794, 990, 1058, 1060, 1296, 1394, 1398, 1400,
1402, 1520, 1522.

\begin{table}[htbp]
  \centering
  \begin{tabular}{|l|c|c|}\hline
Method    & Variables & NMSE (test) \\\hline
PCR       & 8 & $3.64\, 10^{-1}$\\
PLSR      & 4 & $3.67\, 10^{-1}$\\
MI + RBFN & 12 & $4.32\, 10^{-1}$\\
B-Splines + MI + RBFN & 4 & $3.75\, 10^{-1}$\\
B-Splines + MI + LR   & 4 & $3.91\, 10^{-1}$\\\hline
  \end{tabular}
  \caption{Normalized mean squared error on the test set for the cetane number
    prediction problem (diesel database)}
  \label{table:diesel:cn}
\end{table}

Important variables for the PCR linear model form three intervals:
$[974,1086]$, $[1200,1262]$ and $[1486,1550]$. The total range of the
intervals found by the proposed method and by the PCR are of the same order of
magnitude. 

The results of the proposed method are comparable to the ones obtained by PCR,
both in terms of prediction quality and for interpretation purpose: the
wavelength ranges correspond in both cases to absorption bands of hydrocarbons
whose combustibility explains the value of the cetane number. The main
interest of the proposed method, in this case, is to reduce the variable
selection time and to improve the quality of the selected variables compared
to the mutual information based selection of original spectral variables. 

The reasonable running time of the proposed method allows one to test it
regardless of the potential improvements. The fact that the method behaves
similarly as other ones, both in terms of prediction ability and
interpretation, when the setting is \emph{a priori} not advantageous for it,
proves that the method may be used blindly in a wide range of circumstances.

\subsubsection{Discussion}
The results obtained on both data sets show that the computation times for the
selection of variables are drastically reduced compared to a forward-backward
selection procedure carried out directly on the spectral variables. The whole
procedure (B-splines representation, feature selection and nonlinear model
construction) takes a few minutes on a standard personal computer: the
proposed method is therefore very attractive as it can be tested quickly, even
for data sets for which it has no particular reason to outperform linear
methods. This is for example the case when the number of spectral variables is
reasonable and/or when the spectrometric method is known to lead to linear
dependency between the target variable and the spectrum, as in transmission
spectroscopy.

Moreover, the proposed approach reaches similar levels of performances as the
PLSR although it uses the information of far less variables. The variables
that are used are furthermore grouped into consecutive segments, which can
then easily be interpreted.

\section{Conclusion}
Estimating the relevance of spectral variables in a prediction problem is not
an easy task.  The PLSR approach allows scoring each variable by the influence
they have on the PLS components. Unfortunately, a large number of variables
are most of the time taken into account in each PLS component by the method,
making the results difficult to interpret.  An alternative is the selection of
variables with a \emph{wrapper} approach that uses the performances of the
prediction model to score the variables. However, this method demands large
amounts of computations, often rendering the task intractable. Furthermore,
its results may sometimes be difficult to interpret because it may select a
variable and discard another although they may be highly correlated and hence
carry virtually the same information. To overcome these limitations, it is
proposed to gather consecutive variables into groups and to use the
forward-backward selection method to select ranges of frequencies, instead of
selecting individual spectral variables. This is done by means of a B-spline
functional basis to describe the spectra.  Each spectrum is described by a
reduced set of new variables each one related to a range of frequencies.  The
set of new variables being much smaller than the original set of spectral
variables, the computation load is drastically reduced; this renders the
selection procedure feasible even when the spectra contain thousand or more
spectral variables.  Due to the localization properties of the B-splines, the
new variables remains interpretable as they correspond to sub-ranges of the
original wavelength interval. The experiments conducted on spectra obtained
from fescue grass and from diesel fuel show that a nonlinear prediction model
built on the reduced set of variables achieves similar performances as the
PLSR model, although it uses the information from far less variables.  In
addition to reduced computation time and similar (sometimes better)
performances, the method always uses a limited range of spectral variables,
leading to an easy interpretation of the results.

\section{Acknowledgment}
M. Verleysen is Research Director of the Belgian F.N.R.S. (National Fund for
Scientific Research). D. Fran\c{c}ois is funded by a grant from the Belgian
F.R.I.A. Part of this research result from the Belgian Program on
Interuniversity Attraction Poles, initiated by the Belgian Federal Science
Policy Office.  The scientific responsibility rests with its authors. 

The authors thank the anonymous referees for their valuable comments that
helped improving this paper. The authors thank Prof. Robert A. Lodder,
Prof. Fred McLure, and Dr. Barry M. Wise for providing details on the databases
used in this paper. 

\bibliographystyle{elsart-harv}
\bibliography{chimio}

\end{document}